# A Computer-Aided System for Determining the Application Range of a Warfarin Clinical Dosing Algorithm Using Support Vector Machines with a Polynomial Kernel Function

Ashkan Sharabiani, Adam Bress, William Galanter, Rezvan Nazempour, and Houshang Darabi, *Senior Member, IEEE*

*Abstract*— Determining the optimal initial dose for warfarin is a critically important task. Several factors have an impact on the therapeutic dose for individual patients, such as patients' physical attributes (Age, Height, etc.), medication profile, co-morbidities, and metabolic genotypes (CYP2C9 and VKORC1). These wide range factors influencing therapeutic dose, create a complex environment for clinicians to determine the optimal initial dose. Using a sample of 4,237 patients, we have proposed a companion classification model to one of the most popular dosing algorithms (International Warfarin Pharmacogenetics Consortium (IWPC) clinical model), which identifies the appropriate cohort of patients for applying this model. The proposed model functions as a clinical decision support system which assists clinicians in dosing. We have developed a classification model using Support Vector Machines, with a polynomial kernel function to determine if applying the dose prediction model is appropriate for a given patient. The IWPC clinical model will only be used if the patient is classified as "Safe for model". By using the proposed methodology, the dosing model's prediction accuracy increases by 15% in terms of Root Mean Squared Error and 17% in terms of Mean Absolute Error in dose estimates of patients classified as "Safe for model".

## I. INTRODUCTION

Warfarin is one of the most commonly prescribed drugs in the united states [1]. Warfarin is prescribed as an anticoagulant to treat and prevent thromboembolic diseases. Determination of the optimal dose for this drug is quite challenging considering its narrow therapeutic index and the substantial inter-patient variability in dose requirements to attain ideal anticoagulation [2]. This means that mis-dosing (overdosing/under dosing) puts patients at risk of thrombosis, such as deep vein thrombosis or pulmonary emboli for under dosing, and bleeding for overdosing. Moreover, this drug ranks as the principal drug-related cause of adverse effects resulting in hospitalization among the elderly [3]. Warfarin dose is determined based on a blood test called an international normalized ratio (INR), which measures anticoagulation activity [4]. An INR of 2 to 3 is targeted for most indications. If the INR surpasses 3, the patient is at higher risk for bleeding [5]. If the INR falls below 2, the patient is at increased risk for thrombosis [6]. The risk of bleeding or thrombosis with warfarin is highest during the initial months of treatment [2]. There are several factors affecting the activity of warfarin, including age, body size, co-morbidities, and genetic variants in the drug metabolizing enzyme, *CYP2C9,* and the drug target, *VKORC1*. Using these factors, several mathematical models have been developed to predict the optimal starting dose. These mathematical models work as dosing clinical decision support (CDS).

These models have several key differences. First, the models have been derived using the data from patients from a particular geographic and ethnic cohort. Second, some models contain only clinical variables (Clinical models) while others contain both clinical and genetic variables (Pharmacogenetic models). The major limitations of these approaches are that they are based on the availability of genetic data, which is not universally utilized at present. Therefore, in this paper, we concentrated on developing a classification model using only clinical factors.

The proposed model functions as a companion classification model to one of the most popular clinical models in the literature developed by IWPC (International Warfarin Pharmacogenetics Consortium) team known as the IWPC Clinical (*IWPC Cl*) model. Using the multi-ethnic data set collected by IWPC, we have developed a classification model by Support Vector Machines (SVM) with a polynomial kernel function to determine whether *IWPC Cl* model is appropriate for a given patient or not. Hence, the model will get applied for a patient only if the classification model has classified him/her as 'Safe for the model'. The major contribution of this paper is to propose complementary models to one of the most widely used dose prediction model and to determine the correct group of patients for whom the model is well-suited for application.

Several mathematical models are proposed in the literature to assist the clinicians in determining the initial dose. Two linear regression models (one clinical and one Pharmacogenetic) are proposed by *Gage et al.* in 2008. The variables that were applied in the proposed model were BSA

[A].S. PhD was with the Department of Mechanical and Industrial Engineering, University of Illinois at Chicago, Chicago, IL, USA.

A.B. PharmD, MS, is with the Department of Pharmacotherapy, University of Utah, Salt Lake City, UT, USA.

W.G. MS, MD, PhD is with Departments of Medicine, Pharmacy Practice, Pharmacy Systems, Outcomes and Policy, University of Illinois at Chicago, Chicago, IL, USA.

H. D. PhD is with the Department of Mechanical and Industrial Engineering, University of Illinois at Chicago, Chicago, IL, USA. (Corresponding author e-mail: hdarabi@uic.edu).

(Body Surface Area), target INR, Smoking status, Age, and VTE treatment indication [7]. In addition, two linear regression models were developed by the IWPC research team [8]. The data set that was used for modeling is known as the multiethnic data set since it was collected in 9 countries by 21 research teams. The regression models were proposed after investigating several machine learning models, such as regression trees and support vector regression. Despite the model suggested by Gage et al., they used the Age-decade instead of the actual value of Age in their model and claimed that the clinical model is appropriate for patients requiring doses less than or equal to 21 or more than or equal to 49 mg/week. According to 'Clinical Pharmacogenetics Implementation Consortium Guidelines for CYP2C9 and VKORC1 Genotypes and Warfarin Dosing', the models proposed by Gage and IWPC are the most suggested mathematical models in practice [8]. Additionally, several prediction models have been proposed which have used more advanced machine learning methods such as Cosgun et al.[9], Zambon et al. [10], and Sharabiani et al. [11] [12].

Some models were developed to target patients of different ethnicities. For instance, a clinical model was developed by Sharabiani et al. in 2013 for African-American (AA) patients [13]. The proposed model yielded more accurate prediction results than IWPC and Gage models. A Pharmacogenetic model for AA patients was proposed by Hernandez et al. [14]. The developed model outperformed both IWPC models in terms of prediction accuracy. Also, in 2018, Li et al. used back-propagation neural network to create a warfarin maintenance dose prediction model for Chinese patients who have undergone heart valve replacement [15]. Recently, Tao et al. developed an evolutionary ensemble model to improve warfarin dose prediction accuracy for Chinese patients [16]. Moreover, Gaikwad et al. generated a model to predict stable warfarin dose for Indian patients [17]. In addition to the aforementioned models, some models were suggested for children such as Nowak-Göttl et al. [18], Moreau et al. [19], Biss et al. [20], Nguyen et al. [21], and Kato et al. [22].

Although some of the prediction models contained genetic variables, the application of this data is still a controversial issue. Not only does acquiring this data require genetic testing, which limits the applicability range of the models for most clinicians, it is also not guaranteed that involving these genetic variables in the models leads to improvement in clinical endpoints such as time in therapeutic range. Sohrabi and Tajik in 2017, proposed a multi-objective feature approach to select important clinical and genetic characteristics for warfarin dose prediction [23].

In a study known as 'Marshfield Clinic Research Foundation (MCRF)', Burmester et al. investigated the time to reach the therapeutic dose on two patient cohorts. They proved that Pharmacogenetic factors did not accelerate the process of reaching the therapeutic dose [24]. Several research teams in Europe also investigated the impact of applying Pharmacogenetic factors in practice, such as CoumaGen [25], CoumaGen-II [26], Clarification of Optimal Anticoagulation Through Genetics (COAG) [27], and European Pharmacogenetics of Anticoagulant Therapy (EU-PACT) [28]. No robust conclusions were achieved from these studies regarding the involvement of Pharmacogenetic factors on Warfarin dosing. Detailed investigation of the above-mentioned studies are presented in some reviews [29] and [30].

Considering the predominant uncertainty in using the Pharmacogenetic models in practice, in this paper we concentrated on one of the most popular and generally used clinical models; the *IWPC Cl* model. Although, it has been reported that this model performs the best for patients with therapeutic range of less than or equal to 21 to more than or equal to 49 mg/week, since the therapeutic dose is not evident in early stages of the treatment, a companion classification model is proposed to help the clinicians to identify the patients whom are compatible with this dosing model. The proposed model is developed in the framework of support vector machines (SVM) with a polynomial kernel. The remaining sections of this article are as follows: In the next section, first the data set which is used, will be described, then the mathematical background for SVM family models is presented. The modeling and system development procedure will be presented in Section 3. Section 4 will hold the results and finally, conclusion will be presented in Section 5.

II. METHODS

*A. The Data Set*

In this paper, we used the IWPC data set. This data set is publicly available [31] and has been used previously [32]. The data of 4237 patients whose INR was between 2 and 3 were applied for modeling. The imputation method for handling the missing data was to use the average value of the complete cases for the continuous variables and for the categorical variables, using the most-frequent-value of the complete cases. The variables that were available in the dataset are presented in Tables 1,2, and 3 in Appendix Section.

We randomly selected 50% of the data points to the training set (*derivation cohort*) and the remaining 50% to the testing set (*validation cohort*). The points in the training set were applied for developing the classification model. In the preprocessing phase, the binary variables which were significantly unbalanced (one category contained less than 10% of the points) such as Fluvastatin and rifampicin which are now very uncommonly used, and Enzyme (which is short for Enzyme Inducer Status takes the value of 1 if the patient takes carbamazepine, phenytoin, rifampin, and takes the value of 0 the patient does not take any of them) were removed from the data set. Therefore, out of all the variables in the dataset, 13 variables were chosen for the modeling. In the next section, the classification method that is developed in this paper is presented.

*B. Support Vector Machines*

In a classification problem, the data set consists of several features $X_1, X_2, ..., X_L$ and one or several label variables $C_1, C_2, ..., C_p$ representing the classes that the points belong to. The goal is to develop a function to classify the points to their correct classes. Numerous powerful classification models are proposed in Machine Learning literature, such as Decision Trees and Artificial Neural Networks. One of the most powerful classification methods is Support Vector Machine (SVM) which was developed by Vapnik in 1992 [33].

In a binary classification problem ($C_1$ and $C_2$), the objective is to use the $N$ data points in the training set $\{x_n\}_{n=1}^{N}$ to develop a classifier $y(x;w) \simeq w^T\phi(x)+b$ or $y(x;w) \simeq \sum_{i=1}^{M} w_i\phi_i(x)+b$ (1) where $w \in R^M$ is the weight vector, $b \in R$ is the constant and $\phi(.)$ is called the transformation function. The estimated labels are $z_n \in \{-1,1\}$, $n = 1, ..., N$.

Using the *sgn(.)* function; *sgn(y(x))* the points will be assigned to their predicted classes. If the data space is linearly separable, in order to defining the separating hyper plane, the optimal values for $w(w^*)$ and $b(b^*)$ must get estimated. The following definitions are necessary for introducing SVMs.

*Definition 1:* A hyperplane supports a class if it is parallel to a decision surface and all points of its respective class are either above or below. Such hyperplane is known as a supporting hyperplane.

*Definition 2:* The distance between the two supporting hyperplanes is called a margin.

*Definition 3:* A decision surface is optimal if it is equidistant from the two supporting hyperplanes and maximizes their margin.

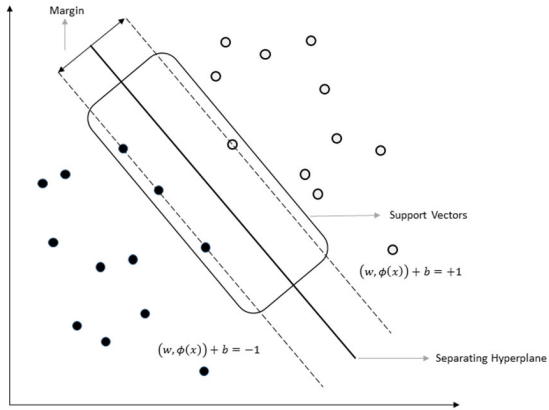

Figure 1. Maximum Margin Classifiers

This optimal hyperplane is called the decision boundary (*D*). The predicted labels for the data points and the value of $y(x_n)$ has the same sign; ( $z_n y(x_n) > 0$; $\forall x_n \in R^D$ and $z_n \in \{-1,1\}$).

The minimum distance of the points in the training set to $D$ is called the *margin* which is computed using $\min_{n\in\{1,...,N\}} \frac{z_n y(x_n)}{\|w\|}$; $\|\cdot\|$ is the $L^2$- norm.

Fig. 1 shows the maximum margin classifiers. The objective in *SVM* is choosing the values for $w$ and b which maximizes the *margin*. The values for $w^*$ and $b^*$ will be yielded by solving the following optimization problem

$$\max_{w \in R^M,\ b \in R} \left\{ \frac{1}{\|w\|} \min_{n\in\{1,...,N\}} [z_n(w^T\phi(x_n)+b)] \right\}. \quad (2)$$

### C. The Kernel Trick

When the data space is not linearly separable, SVMs use a suitable mapping $\langle\Phi\rangle$ of the input data values to a higher dimensional feature space which will be regulated by the kernel function. The data set will be linearly separable in the transformed space. The kernel function returns the inner product of two images of x and x′, i.e., k (x, x′) =⟨Φ(x), Φ(x′)⟩. Based on the nature of the data set, different kernel functions can be most effective: i.e. the polynomial kernel $K(x, x') = (\langle x, x' \rangle + 1)^2$, Multi-Layer Perceptron $K(x, x') = \tanh(\langle x, x' \rangle + \vartheta)$, Gaussian RBF Kernel $K(x, x') = \exp\left(-\frac{\|x-x'\|^2}{2\delta^2}\right)$, ANOVA kernel $K(x, x') = \sum_{k=1}^{n} \text{Exp}(-\sigma(x^k - x'^k)^2)^d$, etc. The kernel function that provided the highest performance in this paper was the polynomial kernel. In the next section the process of selection and training of this model is presented.

### III. MODELLING AND SYSTEM DEVELOPMENT PROCEDURE

#### A. Prediction Model

As mentioned in the previous section, the prediction model which we applied in the system development process is the IWPC clinical model. The variables, their corresponding coefficients, and their units are presented in Table 4.

Table 4. IWPC Clinical Model Coefficients (The predicted dose using this model is the square root of weekly warfarin dose)

| Coefficient | Variable | Unit |
|---|---|---|
| 4.0376 | Intercept | |
| -0.2546 | Age | In decades |
| +0.0118 | Height | In cm |
| +0.0134 | Weight | In kg |
| -0.6752 | Asian | 0/1 |
| +0.406 | Black | 0/1 |
| +0.0443 | Missing | 0/1 |
| +1.2799 | Enzyme | 0/1 |
| -0.5695 | Amiodarone | 0/1 |

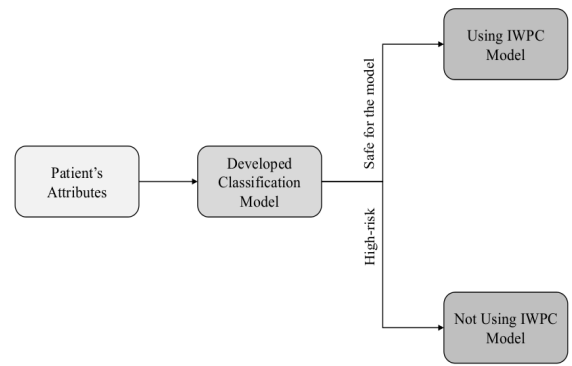

Figure 2. The proposed methodology for using the IWPC clinical model

For all patients in the data set, the dose prediction value using the IWPC model was generated. If the percentage difference between the prediction value and the therapeutic dose (Therapeutic Dose, IWPC Clinical) is more than 15%, the patient will be labeled as 'High-risk' (3252 instances) otherwise he/she will be labeled as 'Safe for the model' (985 instances). The objective is to develop a classification model to detect the High-risk patients. Fig. 2 presents the proposed methodology for classification problem.

For establishing a reliable model and test its performance against the out-of-sample data points, the data set was assigned to Learning (50%) and Testing (50%) sets. The choice of 15% as a threshold was yielded through the consultation with the subject matter experts. Several classification models were examined using K-fold cross validation with k=10 on the learning set. The sensitivity, specificity and accuracy were used in comparing the classification models. The tools used for this project were Python 2.7 (Scikit-Learn 0.18.1) and RapidMiner (7.0).

After developing the model, it can be applied to determine if the patient is compatible with *IWPC Cl* model or not and use the dosing model only if he/she is classified as 'Safe for the model'. After labeling the patients using the classification model, if the patient is classified as "Safe for the model", the clinician has a choice to apply *IWPC Cl*.

### B. Evaluation Measures

There are several methods to evaluate a classification method or a clinical prediction model. We chose to look at the Accuracy, Sensitivity, and Specificity according to the following traditional definitions;

*True positives* (TP): number of positive examples that were predicted correctly

*False positives* (FP): number of negative examples that were predicted incorrectly (Type I error)

*True negatives* (TN): number of negative examples that were predicted correctly

*False negatives* (FN): number of positive examples that were predicted incorrectly (Type II error)

$Accuracy = \frac{TP+TN}{TP+TN+FP+FN}$ (3)

$Sensitivity = \frac{TP}{TP+FN}$ (4)

$Specificity = \frac{TN}{TN+FP}$ (5)

## IV. RESULTS

Several classification methods were examined using the test data set and were compared based on their Accuracy, Sensitivity, and Specificity. The classification methods are Decision Trees (DT) with several parameter settings for minimum size for leaves, depth of the tree and minimum branch size, Artificial Neural networks, SVM with linear kernel, SVM with Gaussian kernel, and SVM with a polynomial kernel. The classification results are presented in Table 5. The SVM with polynomial kernel performed acceptably by having the highest Sensitivity (76.07) and specificity (26.94).

Table 5. Comparing the performance of different classification models on the Test set

| Model | | Accuracy | Sensitivity | Specificity |
|---|---|---|---|---|
| Decision Trees | DT (2,25,4) | 62.84 | 75.06 | 22.24 |
| | DT (2,20,4) | 63.83 | 75.29 | 22.89 |
| | DT (2,30,4) | 62.09 | 74.91 | 21.88 |
| Neural Networks | Neural Nets | 72.14 | 75.8 | 25.8 |
| Support Vector Machines | SVM (Linear) | 68.89 | 75.6 | 23.7 |
| | SVM (Sigmoid) | 69.3 | 75.05 | 18.69 |
| | SVM (Polynomial) | 69.7 | **76.07** | **26.94** |

The SVM with a polynomial Kernel was applied to the patients in the test set to classify patients as either 'Safe for model' or 'High-risk'. Once the patients were classified as 'High-risk', they were eliminated from the test set. For the remaining patients (Shrunken test set), the IWPC clinical model was used to predict the initial dose.

Table 6. Comparing the prediction accuracy of the IWPC CL model on original and shrunken test

| Test Set | Original Test Set | Shrunken Test Set |
|---|---|---|
| Error (RMSE) | 12.66 | 10.82 |
| Error (MAE) | 10.71 | 8.87 |

In Table 6, the prediction accuracy of the IWPC clinical model was compared between the original test set and the shrunken test set based on:

RMSE (Root Mean Squared Error) = $\sqrt{mean[(Actual\ Value - Predicted\ Value)^2]}$ (6)

MAE (Mean Absolute Error) = $mean\ (|Actual\ Value - Predicted\ Value|)$ (7)

After applying the proposed classification of "high risk" or "safe for model", the model's prediction error improved from 12.6 to 10.8 (1.8 absolute, 15% relative) for RMSE and similarly for the MAE method, improved from 10.7 to 8.8 (1.9 absolute, 17% relative). The proportion of patients that would be considered "high-risk" in any new set of patients cannot be determined prospectively and this is something that would need to be watched if this system were to be used on a new cohort of patients.

Clinically the knowledge of whether the patient was classified as "safe for model" or "high-risk" can be used to help decide on the use of clinical pharmacists, which are often a limited resource in healthcare settings. The "high-risk" patients may be the ones that a limited number of pharmacists are assigned to help with anticoagulation. Most patients being started on warfarin do not require continued admission until the INR is stable due to the use of low molecular weight heparin (LMWH). Knowledge of stratification of patients as "high-risk" for a poor dose could potentially be used to help for deciding the delay from discharge to the first visit for ambulatory monitoring of INR.

## V. CONCLUSION

In this paper, a novel methodology for identifying patients appropriate for the IWPC clinical model is proposed, functioning as a companion to IWPC clinical model. The multi-ethnicity (IWPC) data set was used to develop, examine, and ultimately select the best classification model to identify the 'Safe for model' patients; the patients for whom the percentage difference between the prediction by IWPC clinical model and their therapeutic dose is less than 15%, and 'High-risk' patients; the patients for whom the difference between the prediction by IWPC clinical model and their therapeutic dose is more than 15%. A support vector machine with a polynomial kernel function was found to be the best performing classification model. The patients classified as 'High-risk' were eliminated from the test set. For the remaining patients, the IWPC clinical model is used for predicting the initial dose. The performance of the approach was tested using RMSE (Root Mean Squared Error) and MAE (Mean Absolute Error) comparisons on the original test set and the shrunken test set. The RMSE value improved by 15% and the MAE value by 17%. The application of the proposed methodology can be extended to the prediction models which are developed for specific ethnic groups and children.

The objective of this research is not to create a new hybrid predictive model (SVM assisted IWPC model) as a new competing dosing technique to the models mentioned in the Introduction. Our objective mainly is to propose the idea of developing companion classifiers for the dose predication models and identify the appropriate cohort of patients for any dosing model. This idea can serve as a template for other popular dosing techniques. Regardless of the clinician's preference in the dosing technique, a companion classifier can reduce the risk of its application.

The ability of this system to predict which patients may be appropriate or inappropriate for the IWPC model may have many clinical applications. This system could be used to help decide on the use of clinical pharmacists in assistance with warfarin dosing. The "high-risk" patients may be the chosen as requiring pharmacy assistance in a situation with limited clinical pharmacists. In addition, stratification of patients as "high-risk" for a poor dose could potentially be used to help decide the delay from discharge to the first visit for ambulatory monitoring of INR.

APPENDIX

Table 1. Data set description- Categorical variables*

| Var. | V | F | % | Var. | V | F | % |
|---|---|---|---|---|---|---|---|
| Amiodarone | 0 | 3434 | 81 | Gender | 0 | 1822 | 43 |
| | 1 | 228 | 5 | | 1 | 2415 | 57 |
| | M | 575 | 14 | | M | 0 | 0 |
| Aspirin | 0 | 2667 | 63 | Lovastatin | 0 | 2203 | 52 |
| | 1 | 905 | 21 | | 1 | 38 | 1 |
| | M | 665 | 16 | | M | 1996 | 47 |
| Atorvastatin | 0 | 2028 | 48 | Macrolide | 0 | 2227 | 53 |
| | 1 | 233 | 5 | | 1 | 6 | 0 |
| | M | 1976 | 47 | | M | 2004 | 47 |
| Congestive Heart Failure | 0 | 2453 | 58 | Phenytoin | 0 | 2210 | 52 |
| | 1 | 484 | 11 | | 1 | 24 | 1 |
| | M | 1300 | 31 | | M | 2003 | 47 |
| Carbamazepine | 0 | 2210 | 52 | Pravastatin | 0 | 2175 | 51 |
| | 1 | 29 | 1 | | 1 | 66 | 2 |
| | M | 1998 | 47 | | M | 1996 | 47 |
| Current Smoker | 0 | 2554 | 60 | Rifampin | 0 | 2230 | 53 |
| | 1 | 384 | 9 | | 1 | 3 | 0 |
| | M | 1299 | 31 | | M | 2004 | 47 |
| DVT/PE | 0 | 3846 | 91 | Rosuvastatin | 0 | 2220 | 52 |
| | 1 | 391 | 9 | | 1 | 14 | 0 |
| | M | 0 | 0 | | M | 2003 | 47 |
| Diabetes | 0 | 2337 | 55 | Simvastatin | 0 | 3035 | 72 |
| | 1 | 543 | 13 | | 1 | 558 | 13 |
| | M | 1357 | 32 | | M | 644 | 15 |
| Enzyme | 0 | 4150 | 98 | Sulfonamide | 0 | 2223 | 52 |
| | 1 | 87 | 2 | | 1 | 11 | 0 |
| | M | 0 | 0 | | M | 2003 | 47 |
| Fluvastatin | 0 | 2350 | 55 | VR | 0 | 2175 | 51 |
| | 1 | 10 | 0 | | 1 | 645 | 15 |
| | M | 1877 | 44 | | M | 1417 | 33 |

*Var. means variable, V, and F mean the values and frequency of the parameter, respectively. Also, M means the Missing values. DVT/PE is "Deep Vein Thrombosis and Pulmonary Embolism" variable and VR is "Valve Replacement" variable.

Table 2. Data set description- Categorical variables (Age and Race) *

| Variable | Values | Frequency | Percent |
|---|---|---|---|
| Age | 1 | 9 | 0% |
| | 2 | 94 | 2% |
| | 3 | 189 | 4% |
| | 4 | 441 | 10% |
| | 5 | 803 | 19% |
| | 6 | 1020 | 24% |
| | 7 | 1129 | 27% |
| | 8 | 510 | 12% |
| | 9 | 28 | 1% |
| | Missing | 14 | 0% |
| Race | 1 | 2663 | 63% |
| | 2 | 656 | 15% |
| | 3 | 918 | 22% |
| | Missing | 0 | 0% |

* The variable Gender takes 0 for Female patients and 1 for Male patients. The variable Race takes 1 for White, 2 for African-American, and 3 for Asian patients. Consumption of any drug or possession of any disease is indicated with 1 and 0 otherwise. The variable Age is coded in Age-decade format (1 represents 10-19 years old, 2 represents 20-29 etc.).

Table 3. Data set description- Continuous variables

| Parameter/ Variable | Height (Cm) | Weight (Kg) | INR (International Normalized Ratio) | Target INR | Dose |
|---|---|---|---|---|---|
| Missing | 696 | 163 | 0 | 0 | 0 |
| mean | 169.7 | 81.3 | 2.5 | 2.5 | 33.6 |
| Std | 10.6 | 22.7 | 0.3 | 0.1 | 17.4 |
| min | 127 | 34 | 2 | 1.8 | 2.5 |
| max | 202 | 237.7 | 3 | 3.5 | 315 |

CONFLICT OF INTERESTS

None declared.

ACKNOWLEDGMENT

The authors would like to thank Miss Elnaz Douzali for her support and assistance in this project.

REFERENCES

[1]  J. Hirsh, V. Fuster, J. Ansell, and J. L. Halperin, "American Heart Association/American College of Cardiology foundation guide to warfarin therapy," *J. Am. Coll. Cardiol.*, vol. 41, no. 9, pp. 1633–

[2] D. S. Budnitz, M. C. Lovegrove, N. Shehab, and C. L. Richards, "Emergency hospitalizations for adverse drug events in older Americans," *N. Engl. J. Med.*, vol. 365, no. 21, pp. 2002–2012, 2011.
[3] P. Harper, L. Young, and E. Merriman, "Bleeding risk with dabigatran in the frail elderly," *N. Engl. J. Med.*, vol. 366, no. 9, pp. 864–866, 2012.
[4] B. A. Hutten, M. H. Prins, M. Gent, J. Ginsberg, J. G. P. Tijssen, and H. R. Büller, "Incidence of recurrent thromboembolic and bleeding complications among patients with venous thromboembolism in relation to both malignancy and achieved international normalized ratio: a retrospective analysis," *J. Clin. Oncol.*, vol. 18, no. 17, pp. 3078–3083, 2000.
[5] E. M. Hylek, C. Evans-Molina, C. Shea, L. E. Henault, and S. Regan, "CLINICAL PERSPECTIVE," *Circulation*, vol. 115, no. 21, pp. 2689–2696, 2007.
[6] E. M. Hylek et al., "Effect of intensity of oral anticoagulation on stroke severity and mortality in atrial fibrillation," *N. Engl. J. Med.*, vol. 349, no. 11, pp. 1019–1026, 2003.
[7] B. F. Gage et al., "Use of pharmacogenetic and clinical factors to predict the therapeutic dose of warfarin," *Clin. Pharmacol. Ther.*, vol. 84, no. 3, pp. 326–331, 2008.
[8] J. A. Johnson et al., "Clinical Pharmacogenetics Implementation Consortium Guidelines for CYP2C9 and VKORC1 genotypes and warfarin dosing," *Clin. Pharmacol. Ther.*, vol. 90, no. 4, pp. 625–629, 2011.
[9] E. Cosgun, N. A. Limdi, and C. W. Duarte, "High-dimensional pharmacogenetic prediction of a continuous trait using machine learning techniques with application to warfarin dose prediction in African Americans," *Bioinformatics*, vol. 27, no. 10, pp. 1384–1389, 2011.
[10] C.-F. Zambon et al., "VKORC1, CYP2C9 and CYP4F2 genetic-based algorithm for warfarin dosing: an Italian retrospective study," *Pharmacogenomics*, vol. 12, no. 1, pp. 15–25, 2011.
[11] A. Sharabiani, A. Bress, E. Douzali, and H. Darabi, "Revisiting warfarin dosing using machine learning techniques," *Comput. Math. Methods Med.*, vol. 2015, 2015.
[12] A. Sharabiani, E. A. Nutescu, W. L. Galanter, and H. Darabi, "A New Approach towards Minimizing the Risk of Misdosing Warfarin Initiation Doses," vol. 2018, 2018.
[13] A. Sharabiani, H. Darabi, A. Bress, L. Cavallari, E. Nutescu, and K. Drozda, "Machine learning based prediction of warfarin optimal dosing for African American patients," in *2013 IEEE International Conference on Automation Science and Engineering (CASE)*, 2013, pp. 623–628.
[14] W. Hernandez et al., "Ethnicity-specific pharmacogenetics: the case of warfarin in African Americans," *Pharmacogenomics J.*, vol. 14, no. 3, p. 223, 2014.
[15] Q. Li et al., "Warfarin maintenance dose Prediction for Patients undergoing heart valve replacement — a hybrid model with genetic algorithm and Back-Propagation neural network," no. June, pp. 1–11, 2018.
[16] Y. Tao, Y. J. Chen, X. Fu, and B. Jiang, "Evolutionary Ensemble Learning Algorithm to Modeling of Warfarin Dose Prediction for Chinese," *IEEE J. Biomed. Heal. Informatics*, vol. 23, no. 1, pp. 395–406, 2019.
[17] T. Gaikwad, K. Ghosh, P. Avery, F. Kamali, and S. Shetty, "Warfarin Dose Model for the Prediction of Stable Maintenance Dose in Indian Patients," 2018.
[18] U. Nowak-Göttl et al., "In pediatric patients, age has more impact on dosing of vitamin K antagonists than VKORC1 or CYP2C9 genotypes," *Blood*, vol. 116, no. 26, pp. 6101–6105, 2010.
[19] C. Moreau et al., "Vitamin K antagonists in children with heart disease: height and VKORC1 genotype are the main determinants of the warfarin dose requirement," *Blood*, vol. 119, no. 3, pp. 861–867, 2012.
[20] T. T. Biss et al., "VKORC1 and CYP2C9 genotype and patient characteristics explain a large proportion of the variability in warfarin dose requirement among children," *Blood*, vol. 119, no. 3, pp. 868–873, 2012.
[21] N. Nguyen, P. Anley, Y. Y. Margaret, G. Zhang, A. A. Thompson, and L. J. Jennings, "Genetic and clinical determinants influencing warfarin dosing in children with heart disease," *Pediatr. Cardiol.*, vol. 34, no. 4, pp. 984–990, 2013.
[22] Y. Kato et al., "Effect of the VKORC1 genotype on warfarin dose requirements in Japanese pediatric patients," *Drug Metab. Pharmacokinet.*, vol. 26, no. 3, pp. 295–299, 2011.
[23] M. K. Sohrabi and A. Tajik, "Multi-objective feature selection for warfarin dose prediction," *Comput. Biol. Chem.*, vol. 69, pp. 126–133, 2017.
[24] J. K. Burmester et al., "A randomized controlled trial of genotype-based Coumadin initiation," *Genet. Med.*, vol. 13, no. 6, p. 509, 2011.
[25] H. Schelleman et al., "Dosing algorithms to predict warfarin maintenance dose in Caucasians and African Americans," *Clin. Pharmacol. Ther.*, vol. 84, no. 3, pp. 332–339, 2008.
[26] H. Schelleman, N. A. Limdi, and S. E. Kimmel, "Ethnic differences in warfarin maintenance dose requirement and its relationship with genetics," 2008.
[27] B. French et al., "Statistical design of personalized medicine interventions: the Clarification of Optimal Anticoagulation through Genetics (COAG) trial," *Trials*, vol. 11, no. 1, p. 108, 2010.
[28] R. M. F. Van Schie et al., "Genotype-guided dosing of coumarin derivatives: the European pharmacogenetics of anticoagulant therapy (EU-PACT) trial design," *Pharmacogenomics*, vol. 10, no. 10, pp. 1687–1695, 2009.
[29] S. A. Scott and S. A. Lubitz, "Warfarin pharmacogenetic trials: is there a future for pharmacogenetic-guided dosing?," *Pharmacogenomics*, vol. 15, no. 6, pp. 719–722, 2014.
[30] L. H. Cavallari and E. A. Nutescu, "Warfarin pharmacogenetics: to genotype or not to genotype, that is the question," *Clin. Pharmacol. Ther.*, vol. 96, no. 1, pp. 22–24, 2014.
[31] "DataSet." [Online]. Available: https://www.pharmgkb.org/downloads/.
[32] S. M. Öztaner, T. T. Temizel, S. R. Erdem, and M. Özer, "A Bayesian estimation framework for pharmacogenomics driven warfarin dosing: a comparative study," *IEEE J. Biomed. Heal. informatics*, vol. 19, no. 5, pp. 1724–1733, 2015.
[33] V. Vapnik, "Statistical learning theory. Hoboken," *Wiley. Wang, K., Tsung, F.(2007). Run-to-run Process Adjust. using Categ. Obs. J. Qual. Technol.*, vol. 39, no. 4, p. 312, 1998.